\documentclass{midl} %

\usepackage{adjustbox}
\usepackage{tikz}
\renewcommand{\theta}{\vartheta}
\usepackage{placeins} %
\usepackage{float} %

\usepackage{mwe} %
\jmlryear{2022}
\jmlrworkshop{Full Paper -- MIDL 2022}
\usepackage{xcolor}

\midlauthor{\Name{Florentin Bieder\nametag{$^{1}$}} \Email{florentin.bieder@unibas.ch}\\
\Name{Julia Wolleb\nametag{$^{1}$}} \Email{julia.wolleb@unibas.ch}\\
\Name{Robin Sandk\"uhler\nametag{$^{1}$}} \Email{robin.sandkuehler@unibas.ch}\\
\Name{Philippe C. Cattin\nametag{$^{1}$}} \Email{philippe.cattin@unibas.ch}\\
\addr $^{1}$ Department of Biomedical Engineering, University of Basel \\
}

\title{
Position Regression for Unsupervised Anomaly Detection
}
\begin{document}

\maketitle

\begin{abstract}

In recent years, anomaly detection has become an essential field in medical image 
analysis. Most current anomaly detection methods
for medical images are based on image 
reconstruction. In this work, we propose a novel anomaly detection
approach based on coordinate regression. Our method 
estimates the position of patches 
within a volume, and is trained only on data of healthy subjects. 
During inference, we can detect and localize 
anomalies by considering the error of the position estimate of a given patch. 
We apply our method to 3D CT volumes and evaluate it on 
patients with intracranial haemorrhages and cranial fractures.
The results show that our method performs well in detecting these anomalies. 
Furthermore, we show that our method requires less memory than comparable approaches 
that involve image reconstruction. This is highly relevant for 
processing large 3D volumes, for instance, CT or MRI scans.
\end{abstract}


\section{Introduction}
In recent years, anomaly detection has become an essential direction of research 
in medical image analysis. Compared to supervised 
segmentation methods, anomaly detection methods do not rely on 
pixel-wise annotations but on image-level labels instead. This leads
to a much simpler way of annotating the training data and reduces the 
human bias to the model.

We can distinguish two major types of anomaly detection methods 
in the literature: The first type only uses data that is considered 
normal. In terms of medical images, these are images of healthy 
subjects. The second type additionally requires examples of anomalous
data \cite{latentInsensitive, descargan}, and can also be considered as weakly
supervised methods. However, in this work we will focus on the first type
that uses only normal data for the training.

The most widely used methods for image anomaly detection are based 
on reconstruction errors \cite{autoencoderComparison}. These methods 
aim to capture the distribution of the training set of healthy 
subjects by learning a low dimensional representation of the input 
and the reconstruction from this representation back to the original 
image. The core idea is that the correct reconstruction of the input 
will fail in some regions if an anomaly is present in the input image. 
Comparing the output with the input will provide a reconstruction 
error in the pixel space, which can be used as an indicator for anomalies 
\cite{chen2018unsupervised,emergencyCT,autoencoderComparison}.
A challenge in these methods is generating an output
image of high quality and rich in detail. This requirement contributes 
to the computational cost in terms of training time and hardware 
requirements, i.e., GPUs and memory. \cite{tong2021fixingbias} 
have also observed that autoencoders can have a bias towards data 
that can easily be reconstructed and are sensitive to outliers 
in the training set.

This paper presents a novel anomaly detection method for 
medical images based on position regression. In contrast to the 
image reconstruction-based methods our approach operates on 
patches of the input image. Instead of learning a reconstruction 
of the patch in pixel space, our method predicts the location 
of the input patch in the original image. Our method is trained 
only on data of healthy subjects and implicitly learns the 
distribution of the data as a result of the position regression task. 
During inference, a significant error in the position prediction of a 
patch indicates that an anomaly is present within that patch, 
which was not present in the training distribution. A detailed 
overview of our method is shown in \figureref{fig:ppr_overview}.
We evaluate our method on the head CT dataset presented by \cite{cq500},
which contains images of patients with intracranial haemorrhages 
and cranial fractures.

Coordinate regression problems have been explored for 
point-of-interest localization \cite{numericalCoordinateRegression} 
or to propose bounding boxes in object detection 
\cite{girshick2014rich,ren2015faster}. 
However, these methods focus on finding the location of particular 
objects within the input image. Compared to this, our method 
predicts the coordinates of the input patch with respect to the 
remaining part of the source image. 
\cite{contrastivePatch} proposed a method of estimating positions 
of patches to locate specific organs or other anatomical structures within
whole-body CT scans. In contrast to our method, they 
simultaneously feed two patches into
their network and perform a regression over the relative 
position of the two patches.

\section{Method}
The method we propose is based on position regression.
From an input volume image $I \in \mathbb R^{N\times N\times N}$ we extract patches 
$p_{\mathbf{x}} \in \mathbb R^{s_p \times s_p \times s_p}$, with a patch size of $s_p$
at position $\mathbf{x} \in \mathbb R^3$ within $I$.
The voxels in $I$ can be indexed using three dimensional coordinates $(x,y,z)$ with
the coordinates $x,y,z \in \{0,1,2,\ldots, N-1\}$. For our purposes we normalize these coordinates
to the range $[0,1]$ such that $\mathbf x = (x,y,z) \in [0,1]^3$.

\begin{figure}[thbp]
\floatconts
{fig:ppr_overview}
{\caption{Conceptual overview of the our proposed method. }
}{
  \includegraphics[width=0.70\linewidth]{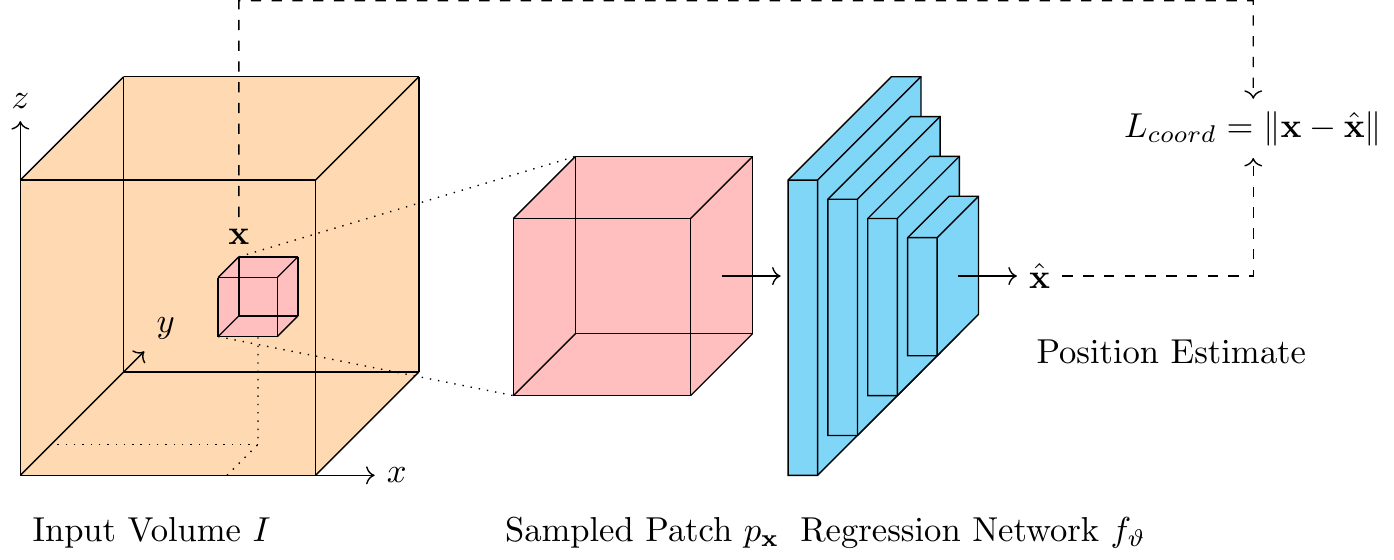}
}
\end{figure}

Using Cartesian coordinates works well for head CT scans, as they
always have the same structures in a similar position. For scans of other anatomical
structures different that may have different poses, other coordinate 
systems might be better suited, for instance, a barycentric coordinate system
based on some key points.

Given some patch, our network $f_\theta: \mathbb R^{s_p \times s_p \times s_p} \to \mathbb R^3$ 
with parameters $\theta$ is trained to 
output an estimate $\hat{\mathbf{x}} = f_\theta(p_{\mathbf{x}})$ of the source position $\mathbf{x}$
of the given patch, as shown in \figureref{fig:ppr_overview}. Thus we 
can consider the training of the network as solving a regression problem. 

For training, we iterate over our training set: In each iteration,
we consider an input volume $I$. We randomly sample coordinates 
$\mathbf{x}$ from the input volume $I$ and extract the surrounding patch 
$p_{\mathbf{x}}$.  Then we pass the patch 
through the network to get the estimated coordinates
$\hat{\mathbf{x}} = f_\theta(p_{\mathbf{x}})$ and use the $\ell^2$-norm of the difference between 
the coordinates of the patch $\mathbf{x}$ and the estimated coordinates $\hat{\mathbf{x}}$ as our
training loss \[
L_{coords}(\mathbf{x}, \hat{\mathbf{x}}) = \Vert \mathbf{x} - \hat{\mathbf{x}}\Vert.
\]

In practice, we use multiple patches in each iteration and average the individual
losses $L_{coords}$ to get our training loss that we optimize. These patches are sampled
independently from a uniform distribution over input coordinates.

During inference, we compute an output volume $E$ of the same size as the input $I$.
For every voxel $I_{\mathbf{x}}$ at coordinates $\mathbf{x}$ we sample the surrounding
patch $p_{\mathbf{x}}$ of the input $I$, and pass it through the network 
to get an estimate $\hat{\mathbf{x}}$. Note that the patches centered at the coordinates
of two neighbouring voxels will overlap.
Then we define the output volume $E$ by computing the reconstruction error for each voxel
as $E_{\mathbf{x}} = L_{coords}(\mathbf{x}, \hat{\mathbf{x}} )$.

This allows us to see the output volume $E$ as an error map. Each $E_{\mathbf{x}}$
shows how well the network $f_\theta$ could predict the position of the patch $p_{\mathbf{x}}$
centered at $\mathbf{x}$.
If the value of a voxel in the output volume $E$ is above a certain threshold, 
the network
failed to predict the correct position of the associated input patch. This is the
case if the input patch exhibits a structure that is not present at that position 
in the training images.
Therefore, we can check for high values in the output volume $E$ to find regions that
appear anomalous.

\subsection{Architecture}
The network we used for the patch position regression (PPR) has 
a generic image 
classification network architecture. 
The detailed 
structure is displayed in \figureref{fig:ppr_architecture}. 
The architecture is parametrized by a network size parameter $m$ to consider
a whole range of networks with a different number of parameters.
This parameter determines the number of channels in the convolutional blocks as well as the
size of the affine layers.
We define two blocks, ``Downsample'' and ``Residual'' that are used throughout
the network. Here ``AvgPool'' stands for average pooling with a kernel of size $2$ and 
stride of size $2$ in every direction. ``Conv($c, k, s$)'' stands for a 3D 
convolution with $c$ output channels, a kernel size of $k$ and a stride of $s$
with spectral normalization. ``Linear($n$)'' is an affine transformation with a codomain 
of dimension $n$.

\begin{figure}[htbp]
\floatconts
{fig:ppr_architecture}
{\caption{Architecture of the Patch Position Regression Network. 
}
}
{
\scalebox{0.8}{
  \raisebox{2.2cm}{
  \scriptsize %
  \begin{tabular}{r|l}
  \# & \textbf{Block} \\ \hline
  1 & Residual($m$) \\
  2 & Downsample($m$) \\
  3 & Residual($2m$) \\
  4 & Downsample($2m$) \\
  5 & Residual($4m$) \\
  6 & Downsample($4m$) \\
  7 & Residual($8m$) \\
  8 & Downsample($8m$) \\
  9 & Residual($16m$) \\
  10 & GlobalAvgPool \\
  11 & Linear($16m$) \\
  12 & LeakyRelu($0.2$) \\
  13 & Linear($3$) \\
  14 & Sigmoid
  \end{tabular}
  }
  \includegraphics[width=0.6\linewidth]{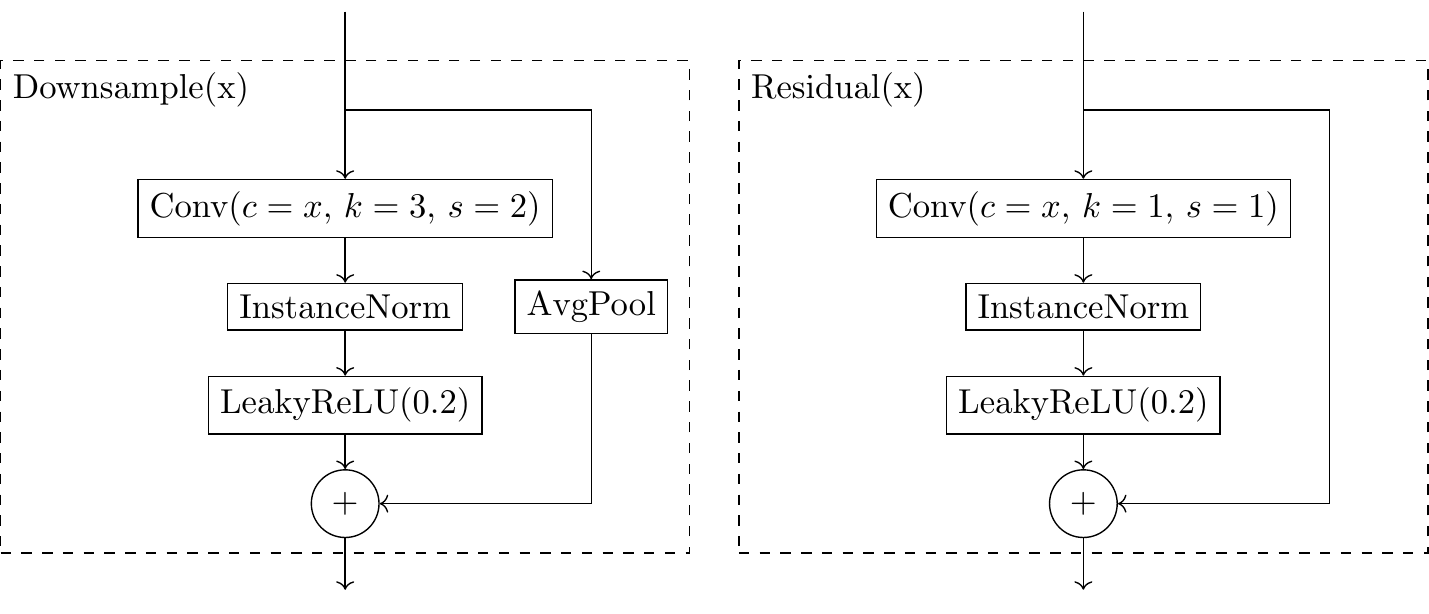}
}
}
\end{figure}

\section{Experiments}\label{sec:experiments}

All runs were trained for 2000 epochs with the Adam optimizer \cite{kingma2014adam} with a learning rate $\text{lr}=10^{-4}$.
We selected the learning rates empirically based on the training 
loss after 200 epochs. 
We manually selected a fixed patch size of $s_p = 31$ (i.e. $31\times 31\times 31$) voxels 
for all experiments. 
(We performed an experiment to examine the influence of the patch size on the performance (Appendix \ref{sec:patch_size})
and found that in this setting the influence is small.)
The size of the patch determines the amount of context: Thus,
the amount of information the network gets,
but also the sensitivity: There is a trade-off between more context,
for a more accurate localization but less sensitivity to anomalies for larger 
patches and a greater sensitivity
but inferior localization accuracy of smaller patches. 

\subsection{Sampling}
In each iteration, we sample 256 patches from one volume. This number
can be adjusted to the memory budget of the given hardware. We set this
number to get a similar run-time as the baseline method (see Section \ref{sec:time}).
Patches that exclusively contain background (i.e. no part of the subject's
anatomy) are discarded for the computation of the loss.

\subsection{Dataset}
For training and evaluation we used the public CQ500 dataset \cite{cq500}, which
contains head CT scans. For some patients, there are multiple
scans present, for instance with and without contrast enhancement. 
Three experts determined for each volume whether there is as an intracranial
haemorrhage (ICH) for each brain hemisphere and whether a cranial fracture is present.
If there was a disagreement between the three raters, we used
the majority vote as our ground truth.
For our experiments, we used one scan from each 
patient without contrast enhancement, 
and discarded all scans that had faulty data (missing slice,
wrong anatomical structure etc). The final dataset used in our experiments contains
131 images without anomaly (111 of which are used for training), and 65 images with
anomalies. The details of the composition of the dataset are
given in Appendix \ref{sec:dataset}.

\subsection{Preprocessing}
The volumes are resampled to voxels of size
$1\times1\times1\mathrm{mm}$ with a total size of $256\times256\times256$ (that is $N=256$) voxels.
Then each volume is rigidly registered to a manually selected reference
volume from the training set using AirLab \cite{sandkuhler2018airlab}.
Since CT images have a high dynamic range, we perform a histogram
equalization.
We segment the skull and the two brain hemispheres in the images. This segmentation
is only used for the evaluation of the method.
Furthermore, we separate foreground from background to be able to filter out irrelevant
patches during training.

\subsection{Autoencoder Baseline}\label{sec:ae_baseline}
We use the basic autoencoder (AE) architecture from \cite{autoencoderComparison} as a baseline
and adapt it to accommodate 3D volumes and to 
the resolution of the volumes in our dataset. The exact architecture is 
show in \figureref{fig:ae_architecture} in Appendix \ref{sec:baseline_architecture}.
As a post-processing step, we applied a filter (suggested in \cite{autoencoderComparison})
of size $5$ to the reconstruction error map.
We used a median filter
for the fractures task, and a grayscale erosion for the ICH task.
We optimized over both filter types and multiple kernel sizes to
make the comparison as fair as possible.

\section{Results}

For comparison, we trained both our proposed method and the baseline 
exclusively on \emph{normal} (healthy)
data. We used the coordinate reconstruction error to detect anomalies: If the error between
the actual and the predicted coordinates is high, we use that as an indication of an
anomalous region. The dataset only includes labels of whether an anomaly
is present in a given structure (e.g. left hemisphere). 
For each of these three structures we check whether the error
exceeds a certain threshold, in order to predict whether an anomaly is present.
Since the performance metrics of the detection, therefore, depends on this threshold,
we report the \emph{receiver operating characteristic}
(ROC) and the corresponding \emph{area under ROC} (AUROC).

\subsection{Computational Resources}
\begin{figure}[thb]
\floatconts
{fig:memory}
{\caption{Performance for multiple values of the network size parameter $m = 2^0, 2^1, \ldots, 2^6$.
Each of the two plots shows the performance on the test set 
as a function of the number of network parameters for the two types of anomalies (ICH, Fracture). The asterisk
marks the best performing PPR model using fewer parameters and less GPU memory than the
best performing AE model (also marked with "$*$").  
}
}
{
\begin{tikzpicture}[scale=0.60, transform shape]
\clip (0,0.6) rectangle (18, 8.7);
\node[anchor=south west] at (0,0) {\includegraphics{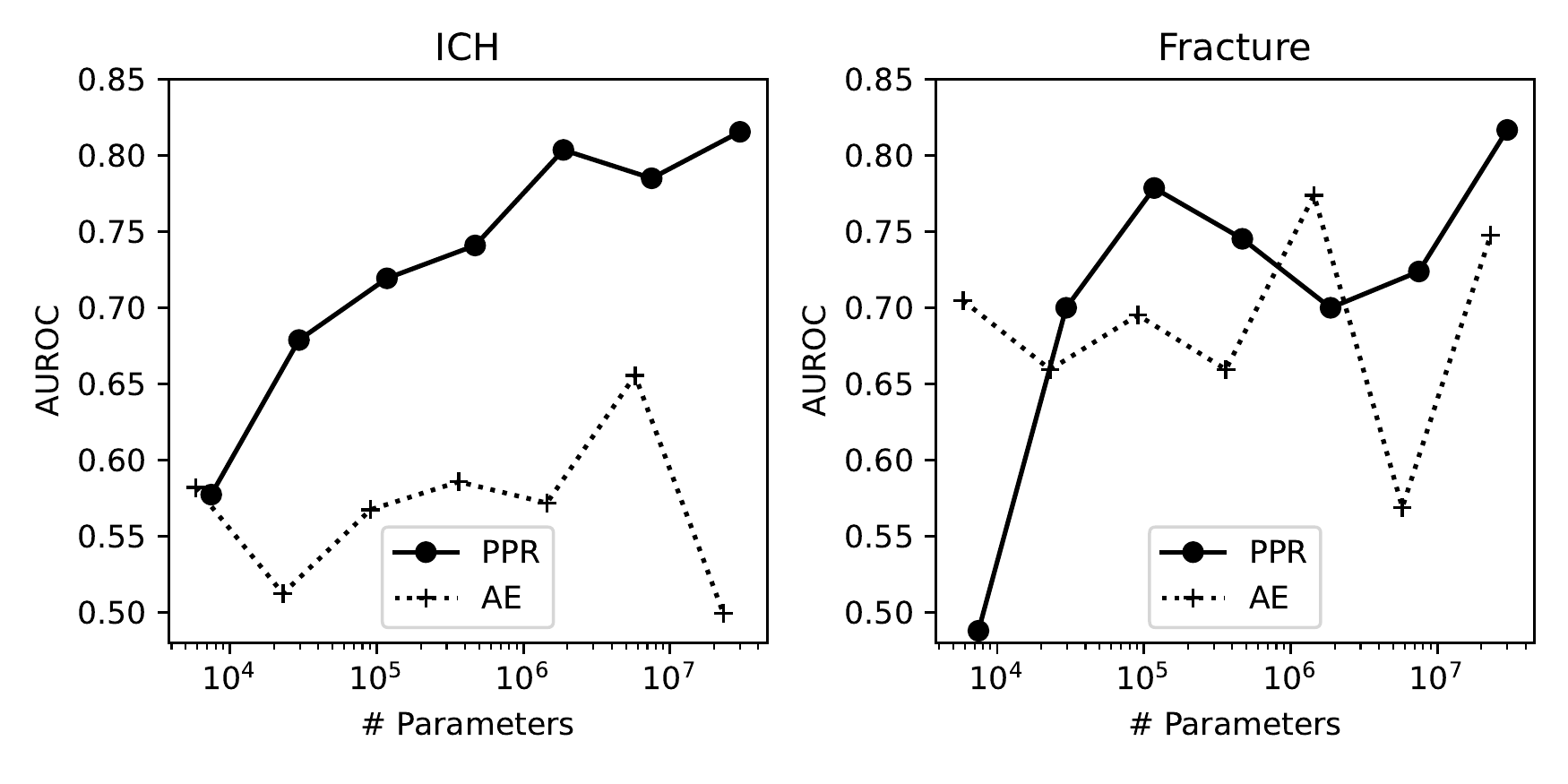}};
\node[anchor=south] at (6.5, 7.3) {\Large *};
\node[anchor=south] at (7.3, 4.8) {\Large *};
\node[anchor=south] at (13.2, 6.9) {\Large *};
\node[anchor=south] at (15.0, 6.8) {\Large *};
\end{tikzpicture}
}
\end{figure}
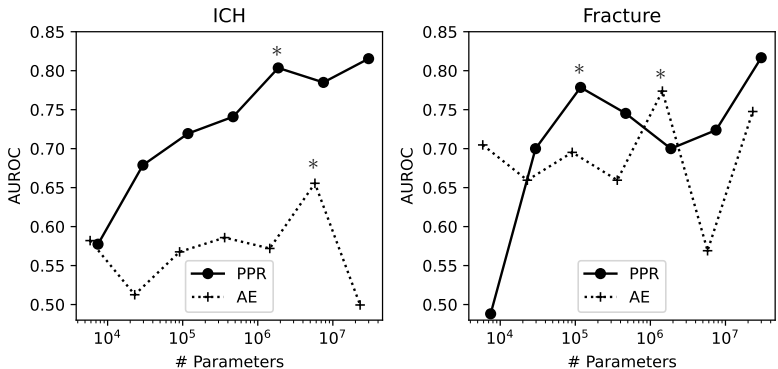

To evaluate the performance of our method and the baseline
method with respect to the computational resources,
we trained both methods with various values of the model size
parameter $m = 2^0, 2^1, \ldots, 2^6$ to 
compare the anomaly detection performance to the size of the network. 
\figureref{fig:memory} shows the performance
as a function of the number of parameters of the networks.
We can see that the performance of the networks increases
up to some limit, but then decreases again. We conjecture that at 
a specific size, the training would benefit from more iterations or
a better initialization.
If we consider the best performing AE networks, we can
see that the PPR network requires roughly one (fractures)
or two (ICH) orders of magnitude
fewer parameters to exceed the performance of the AE.
For the remainder, we discuss the best performing PPR models of those
that both have fewer parameters \emph{and} used less GPU memory 
for the training than the best performing AE models. 
These are marked with an asterisk
in \figureref{fig:memory}.

We want to point out though, that even though the memory requirements
correlate with the number of parameters, they are also
influenced by the actual architecture of the networks as
well as the used software frameworks. Furthermore
the batch size also influences the memory requirements.

\begin{table}[htb]
\floatconts
  {tab:memory}%
  {\caption{GPU Memory requirements (in MB) during training, given some batch size.}}%
  {
\scalebox{1}{
\begin{tabular}{c|rr|rr}
                   &       \multicolumn{2}{c}{ICH} & \multicolumn{2}{c}{Fracture} \\
    Batch Size     & PPR  &    AE &  PPR & AE   \\\hline
 $\text{bs}_{exp}$ & 4452 & 12548 & 1760 & 7894 \\
 $               1$   & 2194 &  6730 & 1000 & 2742 \\
\end{tabular}
}
  }
\end{table}

In our experiments, we used batch sizes that would be used for
practical purposes, that is we have $\text{bs}_{exp}^{\text{PPR}}$  = 256 patches 
(sampled from one volume) for the PPR models and
$\text{bs}_{exp}^{\text{AE}} = 4$ volumes for the AE models. The GPU memory used with these
settings for this is shown in the first row of \tableref{tab:memory},
and we see that our proposed method uses about a factor of 4 less memory.
But even if we only use a batch size of one for the AE models (see second row)
the best performing PPR model still uses a factor of about 1.5 less memory. 
(The memory consumption for all our experiments is shown in
\figureref{fig:network_memory_params} in Appendix \ref{sec:memory_consumption}.)

It should also be noted that in the ICH experiment, there were PPR models that 
outperformed the best performing AE model, and used even less memory during training.
This illustrates, on the one hand, the general issue of the cost
of handling 3D volume data and on the other hand the cost
of the image reconstruction branch of the AE models that is not present in the PPR network.

\subsection{Training and Testing Time requirements}\label{sec:time}
We chose the batch sizes, and number of patches respectively, to result in
a similar training time for all models. The time for the AE models was 
about 26 hours on average, the time for our PPR models is about 22 hours.
While the PPR models require less memory for training,
they are slower than AE models during inference, which is the
price for the patch based processing: The AE models used needed
around $2$ seconds to process one volume, while the PPR models needed
around $1.5$ minutes at the highest resolution.

\subsection{Qualitative Results}
\begin{figure}[htbp]
\floatconts
{fig:qualitative_ich}
{\caption{Slices of some selected examples that show the original CT scan 
with appropriate scaling of the brightness on the left, as well 
as the error map of our proposed
PPR model (with $m = 16$). The images on the left and 
in the center exhibit an anomaly 
(ICH) while those on the right are normal (healthy). 
}
}
{
  \begin{tabular}{@{}c@{ }c@{ }c@{}c@{}}
    \begin{tabular}{@{}c@{}}
      \hspace{2cm} \hphantom{X}ICH \hspace{-2cm} \\
      \includegraphics[width=0.28\linewidth]{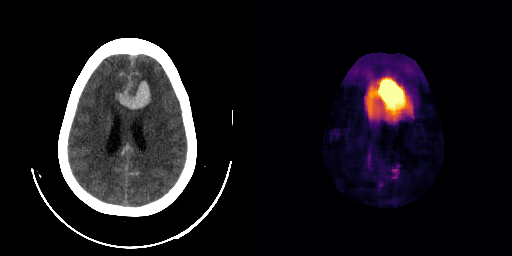} \\
      \includegraphics[width=0.28\linewidth]{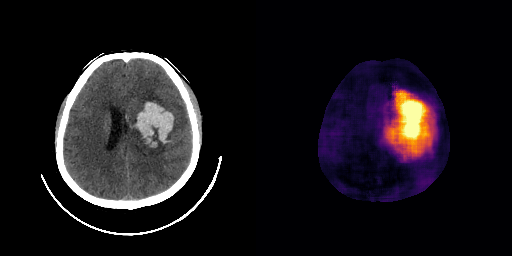} 
    \end{tabular} &
    \begin{tabular}{@{}c@{}}
    \vphantom{X} \\
      \includegraphics[width=0.28\linewidth]{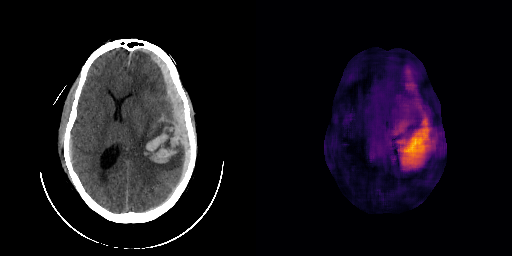} \\
      \includegraphics[width=0.28\linewidth]{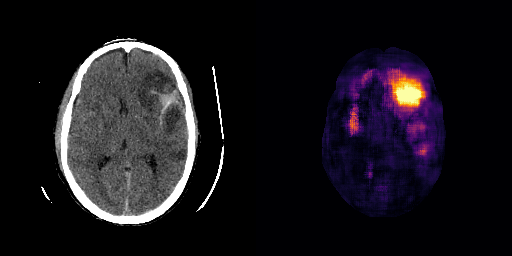} 
    \end{tabular} &
    \begin{tabular}{@{}c@{}}
      healthy \\
      \includegraphics[width=0.28\linewidth]{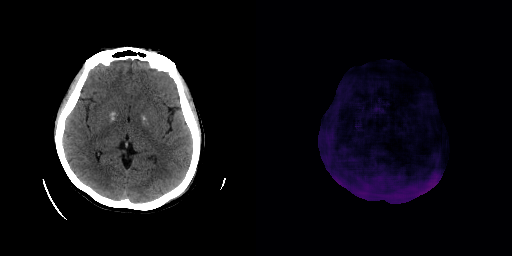}  \\
      \includegraphics[width=0.28\linewidth]{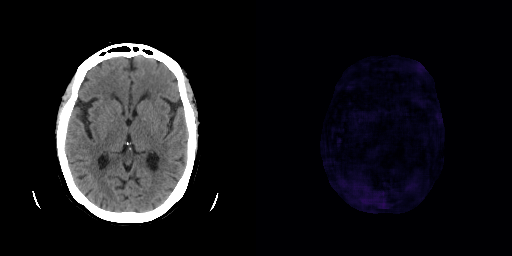}
    \end{tabular} 
  \end{tabular}
  \adjustbox{valign=M}{ %
  \includegraphics[trim={6.0cm 2.2cm 0.5cm 0.3cm}, clip, scale=0.7]{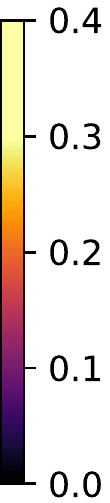}
  }
}
\end{figure}

We show some slices with examples of ICH
in \figureref{fig:qualitative_ich} as well as surface renderings of scans 
of subjects with cranial fractures in \figureref{fig:qualitative_fractures}. (As a reference we also included the same examples for the AE in Appendix \ref{sec:qualitative_ae} in \figureref{fig:qualitative_ich_ae}.)
It is noticeable that the reconstruction error is high where there is
an anomaly. The reconstruction
error generally seems to continuously depend on the amount of the patch that is anomalous, as
the error maps generally seem to be rather smooth. %

The performance for the models used for \figureref{fig:qualitative_ich} 
and \ref{fig:qualitative_fractures} are shown in \figureref{fig:roc_performance}.
We observe see that the detection of fractures is the more challenging
task for our method than the detection of ICH. This might be due to the
smaller number of scans available to evaluate it on (see Appendix \ref{sec:dataset}).
To put these results in context we provided a table with
the inter rater agreement on these tasks in Appendix \ref{sec:interrater}:
The performance in terms of AUROC is around $15-16\%$ lower than the average
raters.

\begin{figure}[htbp]
\floatconts
{fig:qualitative_fractures}
{\caption{Surface renderings of scans of two subjects with cranial fractures.
Both subjects suffer from from fractures of the frontal bone. The left side each shows the scanned
part of the skull and the right part shows the same surface coloured according to the error.}
}
{
  \includegraphics[trim={2cm 4.0cm 10.0cm 3.9cm}, clip, height=0.20\linewidth]
  {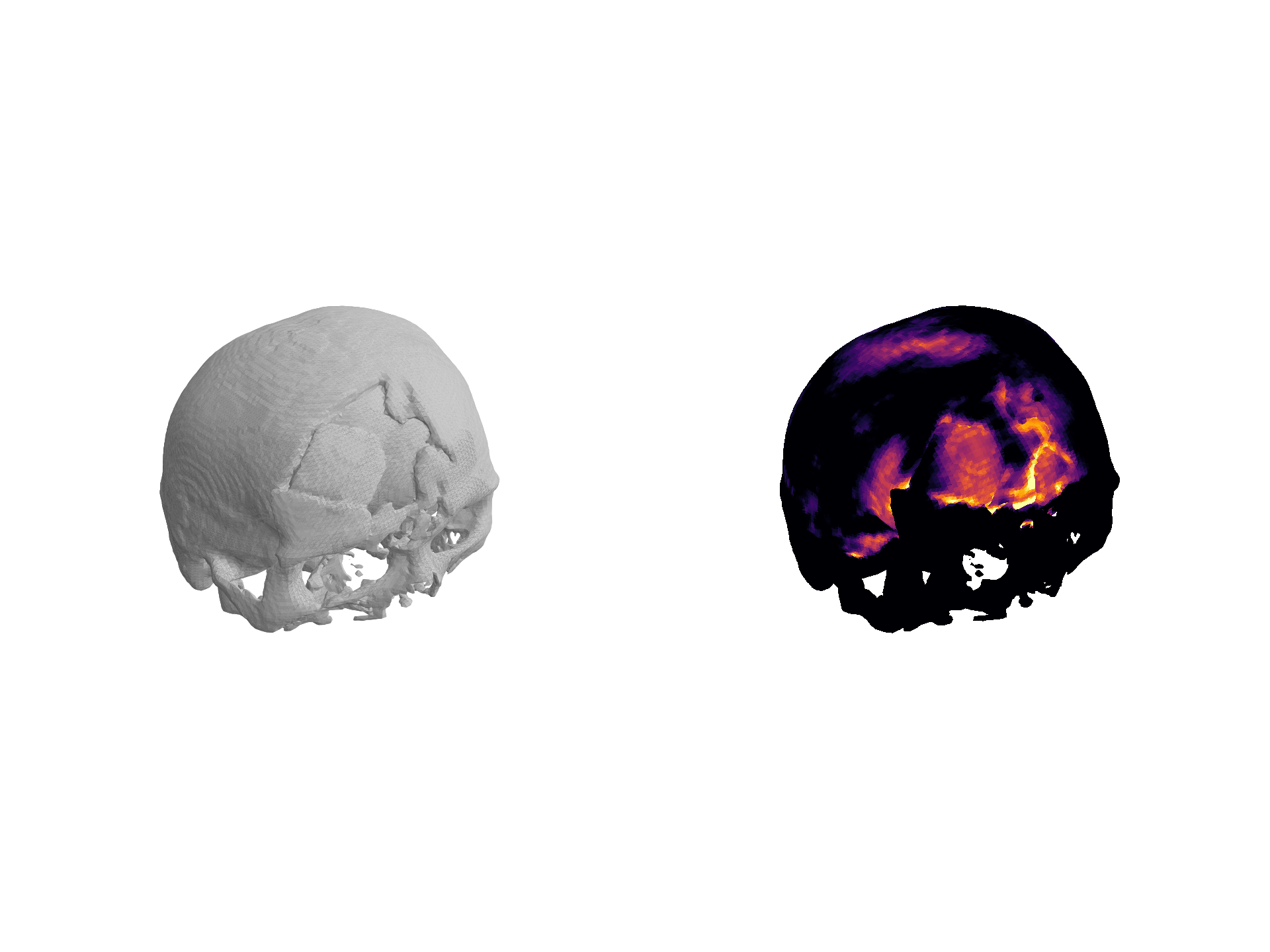}
  \includegraphics[trim={9.5cm 4.0cm 2cm 3.9cm}, clip, height=0.20\linewidth]
  {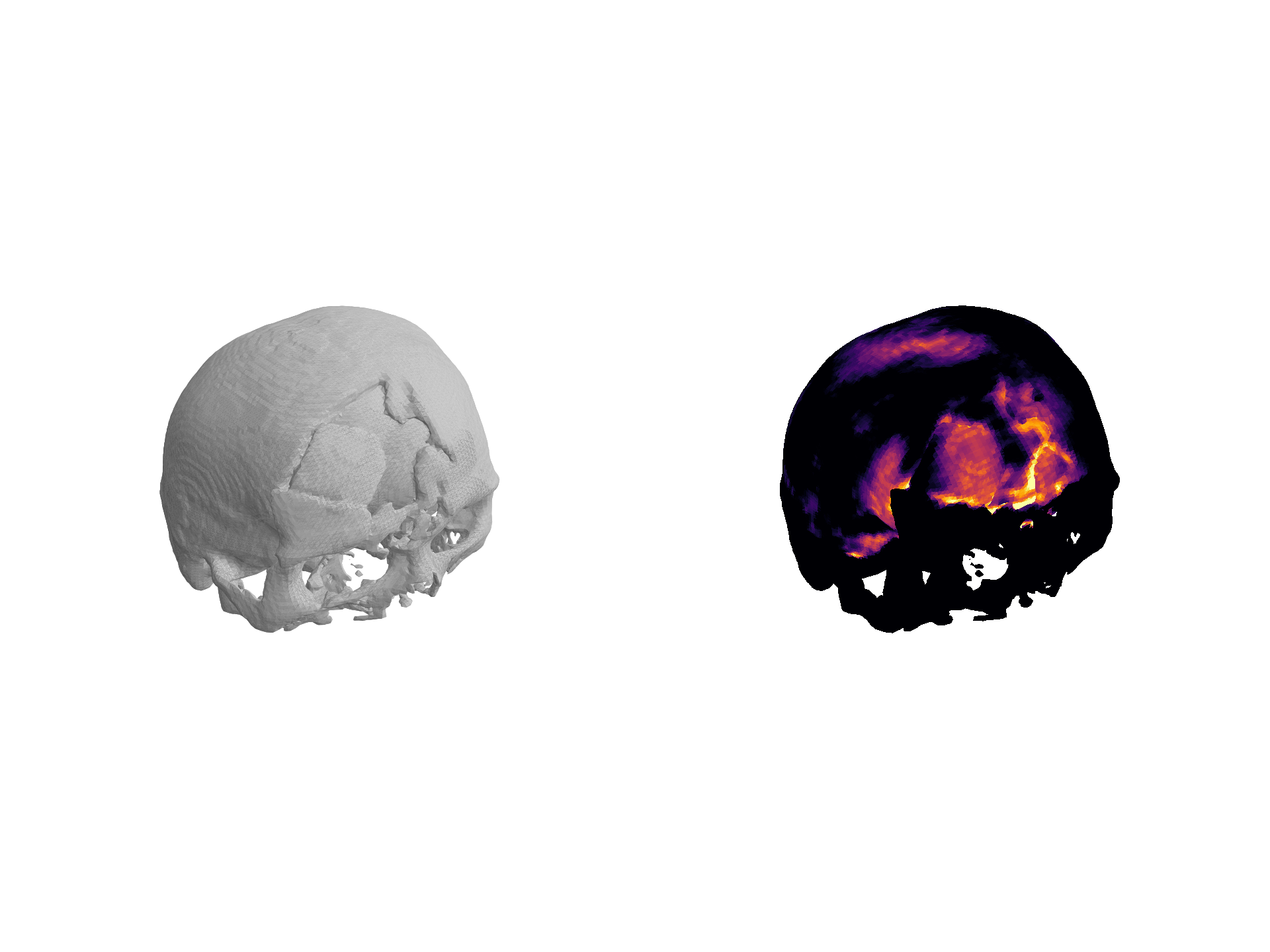}
  \hphantom{1}
  \includegraphics[trim={1cm 4.3cm 9.8cm 4.0cm}, clip, height=0.20\linewidth]
  {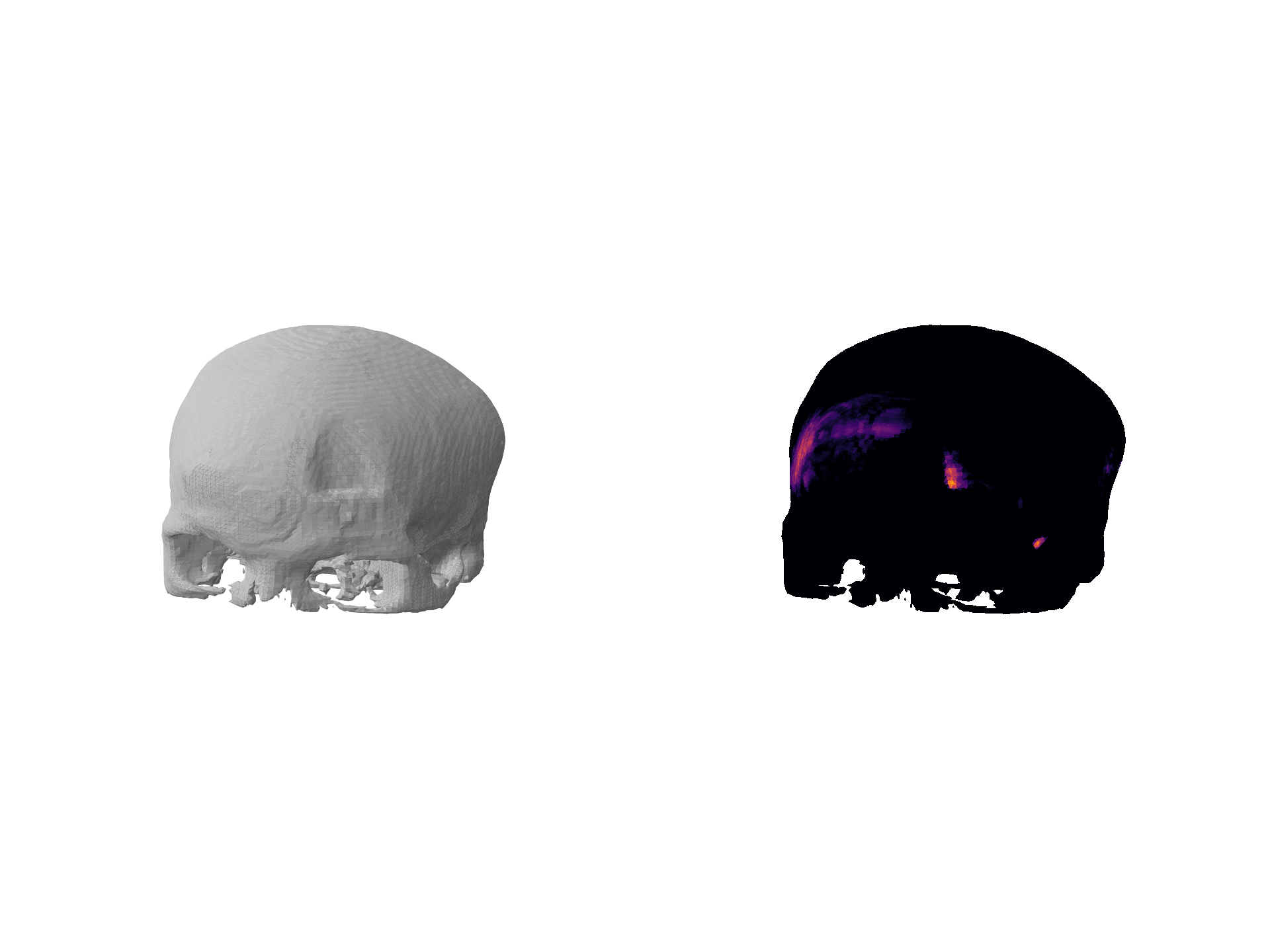}
  \includegraphics[trim={9.5cm 4.3cm 1.8cm 4.0cm}, clip, height=0.20\linewidth]
  {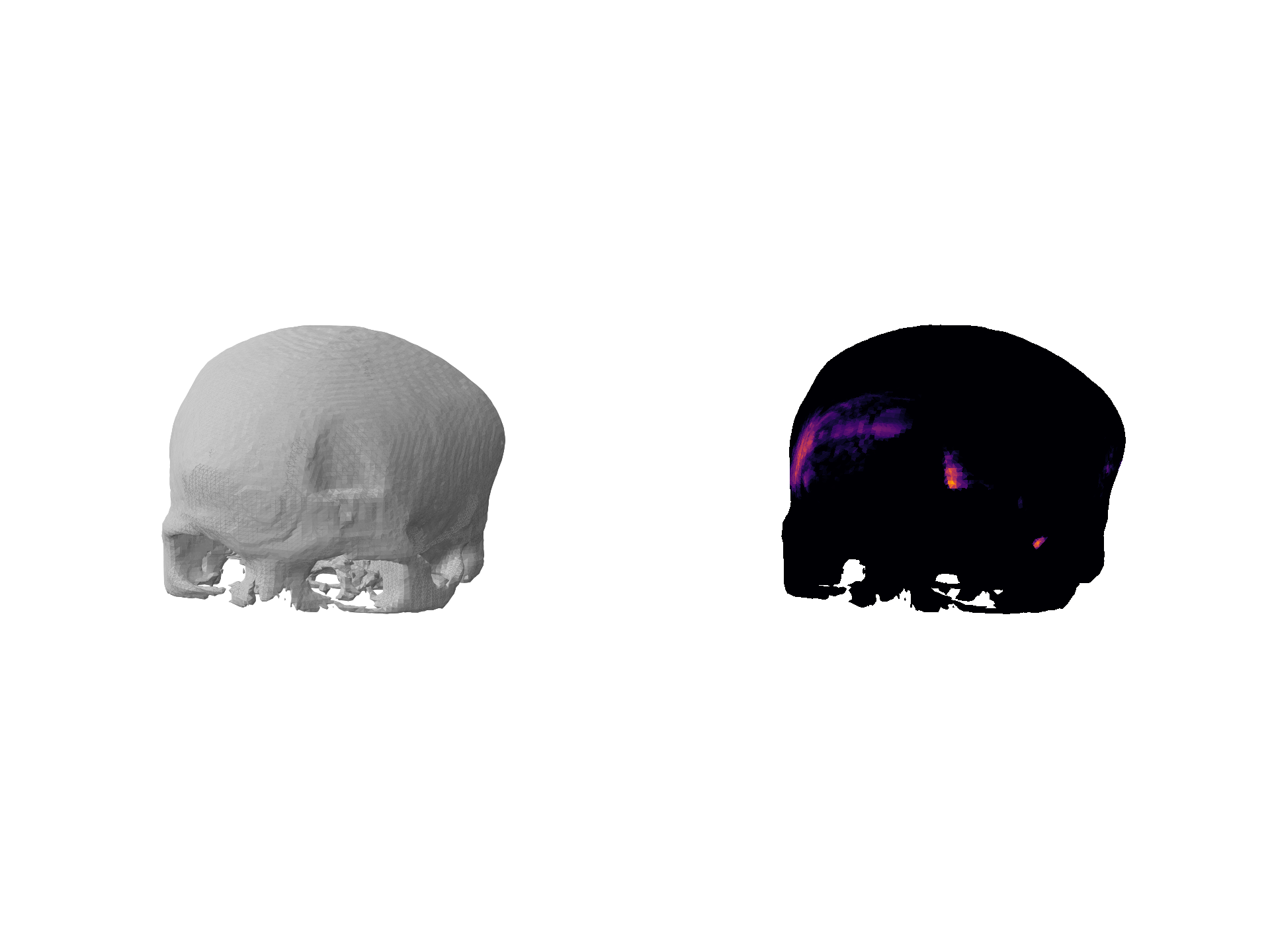}
  
  \includegraphics[trim={0.3cm 0.5cm 0.3cm 6cm}, clip, scale=0.6]{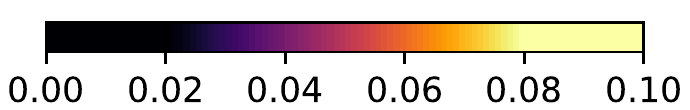}
}
\end{figure}

\begin{figure}[thb]
\floatconts
{fig:roc_performance}
{\caption{
Left: ROC curve for the detection of ICH ($\text{AUROC}=0.81$),
Right: ROC curve for cranial fractures ($\text{AUROC}=0.79$).
}
}
{
  \includegraphics[trim={0 0.45cm 0 0.8cm}, clip, width=0.35\linewidth]{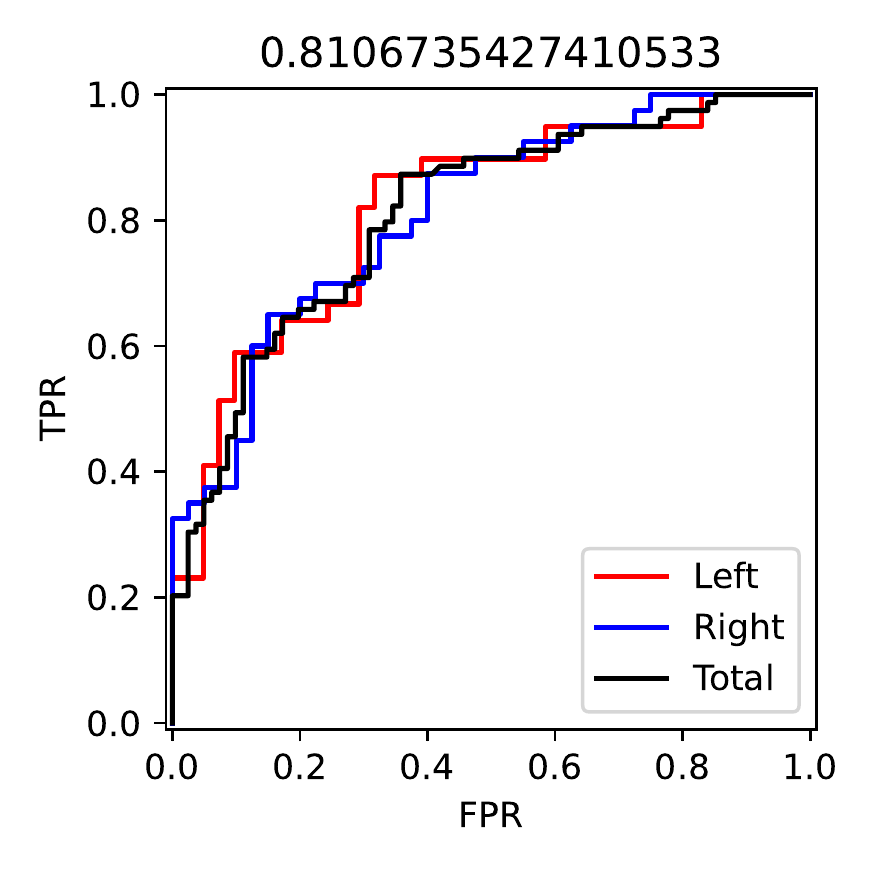}
  \includegraphics[trim={0 0.45cm 0 0.8cm}, clip, width=0.35\linewidth]{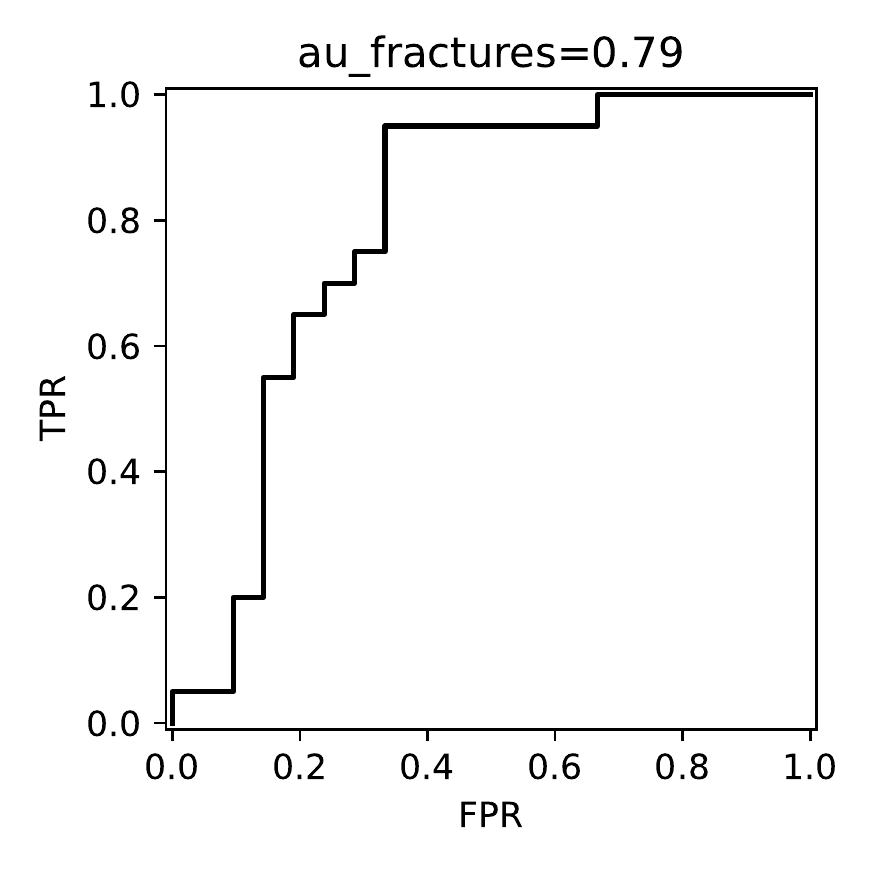}
}
\end{figure}

\section{Discussion}
We have shown that with our proposed approach, we can get a similar
performance to AE models with a memory footprint that is up to an order 
of magnitude lower. This is especially interesting for processing volume
data that are common in medical applications, for instance CT- or MRT scans.
The lower memory requirements come at a price of a longer inference time.
The time needed is still low enough for most applications in the medical
field.

We see the proposed network as a proof-of-concept that would be interesting
for further investigation. One of the drawbacks of this method is the limited
spatial resolution. This could be addressed with a multi stage coarse-to-fine
approach similar to what was proposed in \cite{contrastivePatch}.

Furthermore, it would be interesting to investigate the influence of the
patch size in relation to the spatial extent of the anomalies,
and to see if it would be possible to combine it with 
a tissue classification task.
In addition to the patch size it would also be  to vary
the shape of the patch that is passed into the network

\midlacknowledgments{
We are grateful for the support of the Novartis FreeNovation
initiative and the Uniscientia Foundation (project \#147-2018).
We would also like to thank the NVIDIA Corporation for the
donation of a GPU that was used for our experiments.
}

\newpage

\bibliography{bieder22}

\appendix %

\section{Patch Size}\label{sec:patch_size}
To examine the influence of the patch size (see Section \ref{sec:experiments}) we
evaluated the ICH task with identical settings but different patch sizes. 
For each of the patch sizes we trained a model from scratch with the 
same configuration as in the other experiments.
In \figureref{fig:performance_vs_patch_size} we show the AUROC score
as a function of the patch size for the detection of the 
haemorrhages in each brain hemisphere (red and blue) as well as the
total score for both hemispheres combined.
Across all six patch sizes the performance slightly changes by 
about $\pm 0.06$. Comparing that value with the variation of
the individual brain hemispheres we conclude that 
the influence of the patch size in these experiments in negligible.
For other anomalies or modalities this might be different though.

\begin{figure}[h]
\floatconts
{fig:performance_vs_patch_size}
{\caption{
Performance of the PPR network in terms of AUROC as a function of the patch size
$s_p = 19, 23, 27, 31, 35, 39$ for the ICH task.
}
}
{
\includegraphics[scale=0.7]{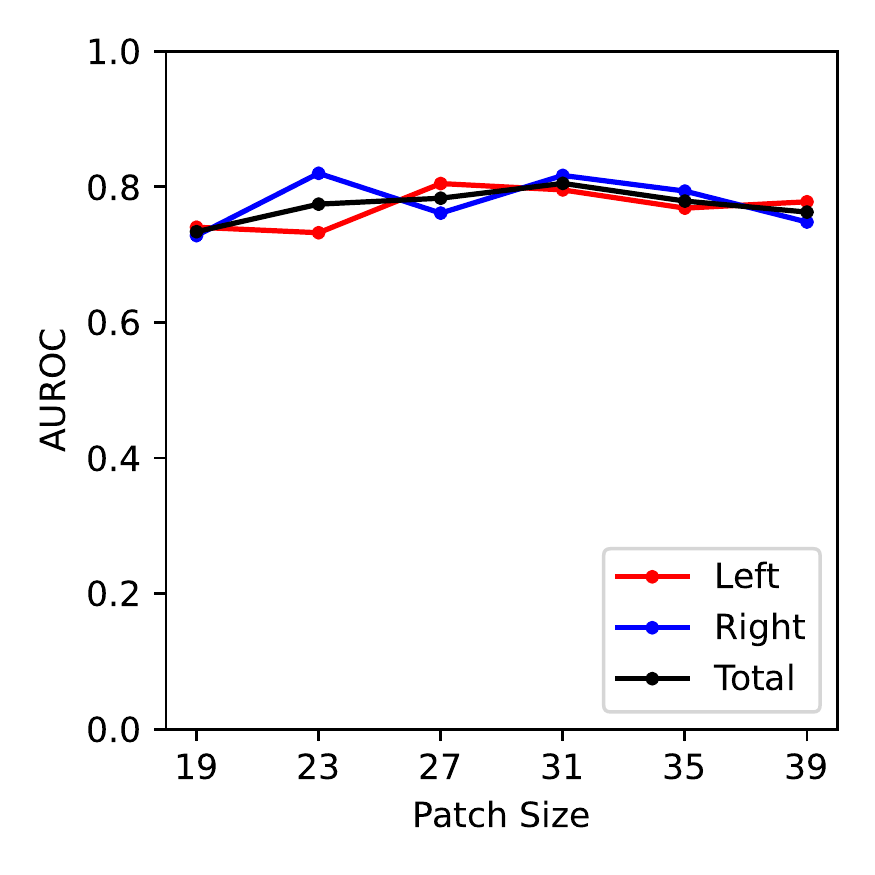}
}
\end{figure}

\section{CQ500 Dataset}\label{sec:interrater}\label{sec:dataset}
The scans with bleeding or a fracture were considered \emph{anomalous}, while
the scans without any of these findings were considered \emph{healthy}.
The test set contains 86 volumes in total, including 21 from the healthy set (i.e. without anomalies), 20 with a fracture, 39 with bleeding in the left
hemisphere and 47 with bleeding in the right hemisphere. Note that these sets are not disjoint,
\tableref{tab:datasplit} shows the actual distribution.
Our test set consists of the anomalous scans as well as 21 
randomly chosen scans from the healthy set. The remaining scans of the healthy
set were used for training. 

\begin{table}[htbp]
\floatconts
  {tab:datasplit}%
  {\caption{Distribution of the anomalous data in the test set. 
  Overlined column/row names indicate the absence of the given feature. Each entry shows the number of volumes with that combination of the presence/absence
  of these three features. 
  }}%
  {
\begin{tabular}{r|rr|rr}
                       &       \multicolumn{2}{c}{ Bleeding Left } & \multicolumn{2}{c}{ $\overline{\text{Bleeding Left}}$ } \\
                       &     Bleeding Right & $\overline{\text{Bleeding Right}}$ &  Bleeding Right & $\overline{\text{Bleeding Right}}$ \\\hline
                     Fracture &      4 &      3 &      7 &      6 \\
 $\overline{\text{Fracture}}$ &     16 &     16 &     13 &     21 \\
\end{tabular}
  }
\end{table}

To characterize the variability of the ground truth within the three raters, we compute the 
Fleiss-Kappa as well as the pairwise AUROC in \tableref{tab:interrater}.  

\begin{table}[htbp]
\floatconts
  {tab:interrater}%
  {\caption{Inter rater agreement expressed using the Fleiss-Kappa as well
  as AUROC for each rater compared to the majority vote for each feature.
  }}%
  {
\begin{tabular}{l|c|ccc|c}
 & $\kappa$ & \multicolumn{4}{|c}{AUROC} \\ 
 &  & R1 & R2 & R2 & AVG \\ \hline 
Bleeding Left & 0.745 & 0.877 & 0.982 & 0.985 &  0.948 \\ 
Bleeding Right & 0.705 & 0.893 & 0.945 & 0.964 &  0.934 \\ 
Fracture & 0.632 & 0.901 & 0.955 & 0.921 &  0.926 \\ 
\end{tabular}
}
\end{table}

\section{Autoencoder Baseline}\label{sec:baseline_architecture}

We trained the AE network for 2000 epochs with the Adam
optimizer with $\text{lr}=0.001$. The architecture of the AE is shown in
\figureref{fig:ae_architecture}: The architecture is parametrized by $m$ to be able to consider networks of different sizes.
We define two blocks, ``Downsample'' and ``Upsample'', that are used throughout
the network.  ``[Transp]Conv($c, k, s$)'' stands for a 3D 
[transposed] convolution with $c$ output channels, a kernel size of $k$ and a stride of $s$. 
The AE models were trained using a pixel-wise $L_1$-loss.

\begin{figure}[ht]
\floatconts
{fig:ae_architecture}
{\caption{Architecture of the AE network. 
}
}
{
  \raisebox{1.7cm}{
  \footnotesize
  \begin{tabular}{r|l}
  \# & \textbf{Block} \\ \hline
  1 & Downsample($m$) \\
  2 & Downsample($2m$) \\
  3 & Downsample($4m$) \\
  4 & Downsample($8m$) \\
  5 & Downsample($16m$) \\
  6 & Upsample($8m$) \\
  7 & Upsample($4m$) \\
  8 & Upsample($2m$) \\
  9 & TranspConv($c=1,k=4,s=2$) \\
  10 & Sigmoid
  \end{tabular}
  }
  \includegraphics[width=0.5\linewidth]{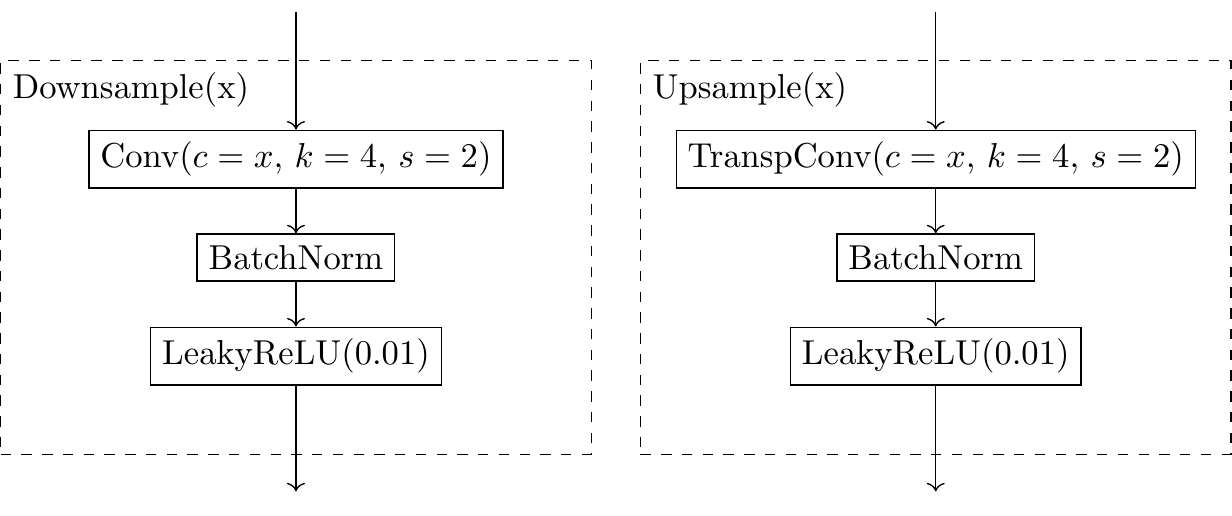}
}
\end{figure}

\newpage

\section{Qualitative Results AE}\label{sec:qualitative_ae}

\begin{figure}[htbp]
\floatconts
{fig:qualitative_ich_ae}
{\caption{Slices of some selected examples that show the original CT scan 
with appropriate scaling of the brightness on the left, as well 
as the error map the baseline AE method. The images on the left and 
in the center exhibit an anomaly 
(ICH) while those on the right are normal (healthy). 
}
}
{
  \begin{tabular}{@{}c@{ }c@{ }c@{}c@{}}
    \begin{tabular}{@{}c@{}}
      \hspace{2cm} \hphantom{X}ICH \hspace{-2cm} \\
      \includegraphics[width=0.28\linewidth]{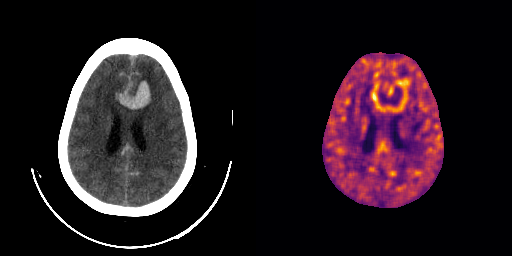} \\
      \includegraphics[width=0.28\linewidth]{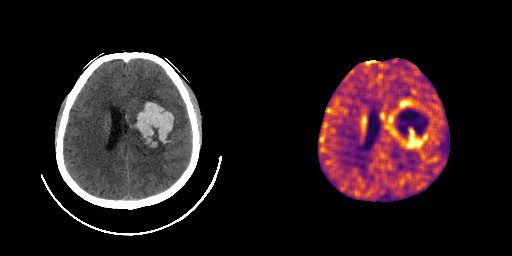} 
    \end{tabular} &
    \begin{tabular}{@{}c@{}}
    \vphantom{X} \\
      \includegraphics[width=0.28\linewidth]{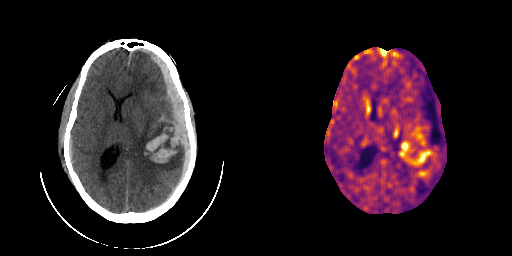} \\
      \includegraphics[width=0.28\linewidth]{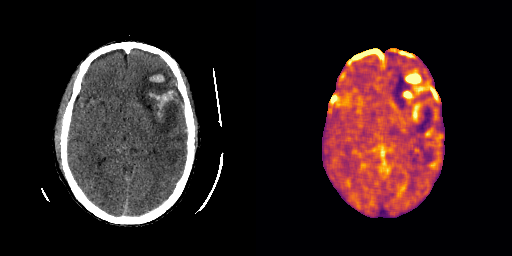} 
    \end{tabular} &
    \begin{tabular}{@{}c@{}}
      healthy \\
      \includegraphics[width=0.28\linewidth]{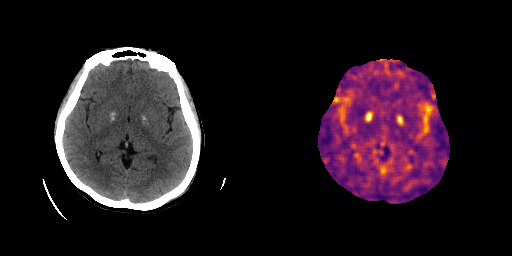}  \\
      \includegraphics[width=0.28\linewidth]{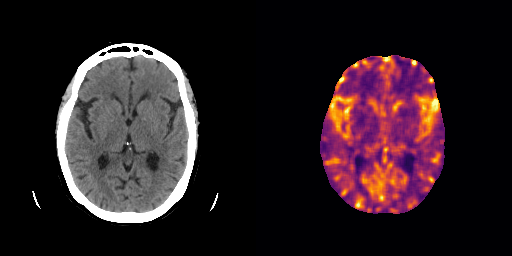}
    \end{tabular} 
  \end{tabular}
  \adjustbox{valign=M}{ %
  \includegraphics[trim={6.0cm 2.2cm 0.4cm 0.3cm}, clip, scale=0.7]{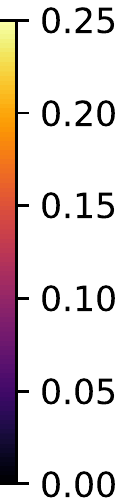}
  }
}
\end{figure}

\clearpage
\section{Memory Consumption}\label{sec:memory_consumption}
\begin{figure}[thbp]
\floatconts
{fig:network_memory_params}
{\caption{GPU Memory consumption as a function of the number of parameters for the AE and PPR with network size parameter $m = 2^0, 2^1, \ldots, 2^6$ and batch sizes $\operatorname{bs}_{exp}$ as used in the experiments.}
}{
  \includegraphics[width=0.35\linewidth]{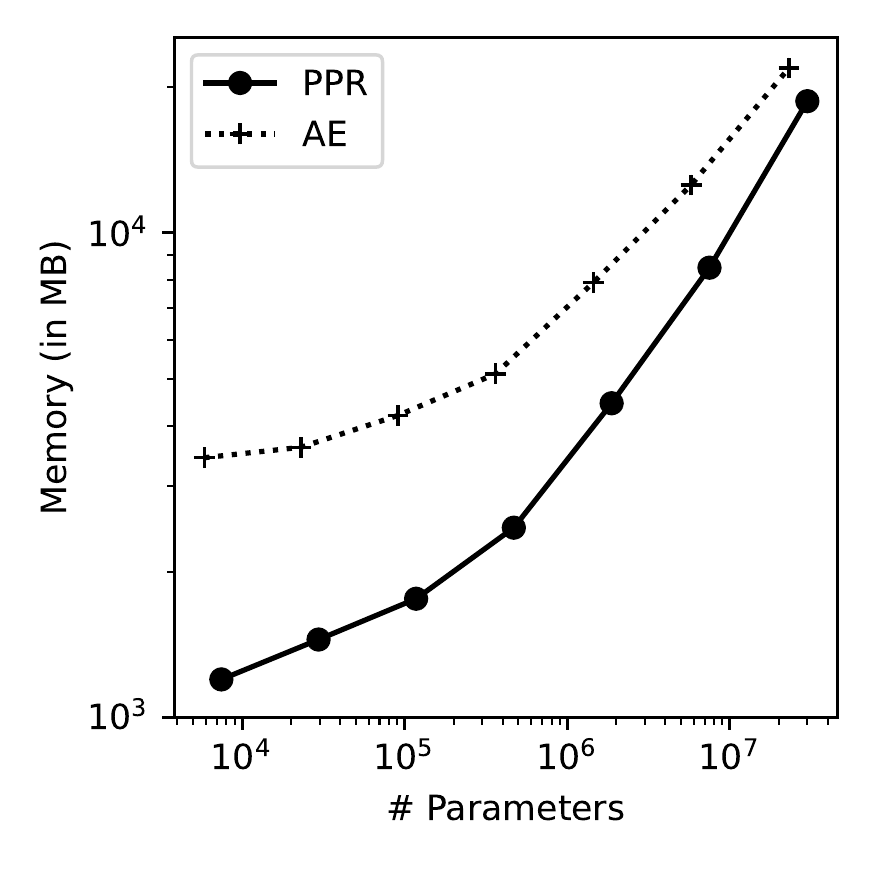}
}
\end{figure}

\end{document}